\documentclass[letterpaper, 10 pt, conference]{ieeeconf}  
\IEEEoverridecommandlockouts                              
\makeatletter
\let\NAT@parse\undefined
\makeatother

\usepackage[table, x11names]{xcolor}
\usepackage[colorlinks=true,linkcolor=blue,citecolor=blue,urlcolor=blue,pagebackref]{hyperref}
\usepackage{times} 
\usepackage{amsmath} 
\usepackage{amssymb}  
\usepackage{graphicx}
\usepackage{algorithm}
\usepackage{mathtools}
\usepackage{algorithm}
\usepackage{algorithmicx}
\usepackage[noend]{algpseudocode}
\usepackage[font=footnotesize]{caption}
\usepackage[font=footnotesize]{subcaption}

\def\figref#1{Fig.~\ref{#1}}
\def\tabref#1{Tab.~\ref{#1}}
\def\eqref#1{Eq.~(\ref{#1})}
\def\algref#1{Alg.~\ref{#1}}

\usepackage{amsopn}

\newcommand{\bt}{\mathbf{t}}
\newcommand{\bM}{\mathbf{M}}
\newcommand{\bA}{\mathbf{A}}
\newcommand{\bB}{\mathbf{B}}

\newcommand{\bC}{\mathbf{C}}
\newcommand{\bH}{\mathbf{H}}
\newcommand{\bI}{\mathbf{I}}

\newcommand{\bX}{\mathbf{X}}
\newcommand{\bZ}{\mathbf{Z}}
\newcommand{\bR}{\mathbf{R}}

\newcommand{\bT}{\mathbf{T}} 

\newcommand{\bJ}{\mathbf{J}}
\newcommand{\bZero}{\mathbf{0}}

\newcommand{\bb}{\mathbf{b}}
\newcommand{\bc}{\mathbf{c}}
\newcommand{\be}{\mathbf{e}}

\newcommand{\bx}{\mathbf{x}}

\newcommand{\br}{\mathbf{r}}
\newcommand{\bz}{\mathbf{z}}

\newcommand{\bh}{\mathbf{h}}

\newcommand{\bDeltax}{\mathbf{\Delta x}}
\newcommand{\bDeltaX}{\mathbf{\Delta X}}

\newcommand{\bDeltam}{\mathbf{\Delta m}}

\newcommand{\bOmega}{\mathbf{\Omega}}

\newcommand{\pluseq}{\mathrel{+}=} 

\DeclareMathOperator*{\argmin}{argmin}

\newcommand{\cN}{\mathcal{N}}

\newcommand{\bXlin}{\breve{\bX}}
\newcommand{\chart}{\mathrm{chart}}

\newcommand{\bZhat}{\hat{\bZ}}
\def\skew#1{{\lfloor{#1}\rfloor}_\times}
\def\flatten#1{\text{flatten}\left({#1}\right)} 

\def\g2o{$g^2o$}

\title{\LARGE \bf Matrix Difference in Pose-Graph Optimization}

\author{Irvin Aloise and Giorgio Grisetti
    \thanks{All authors are with the Department of Computer, Control, and 
    Management Engineering  Antonio Ruberti, Sapienza University of Rome, Rome, 
    Italy, Email: {\tt\small ialoise@diag.uniroma1.it}, {\tt\small 
    grisetti@diag.uniroma1.it}}%
}

\begin{document}
\maketitle
\thispagestyle{empty}
\pagestyle{empty}

\begin{abstract}
  Pose-Graph optimization is a crucial component of many modern SLAM
  systems.  Most prominent state of the art systems address this problem by 
  iterative  non-linear least squares.  
  Both number of iterations and convergence
  basin of these approaches depend on the error functions used to
  describe the problem.  
  The smoother and more convex the error function with respect
  to perturbations of the state variables, the better the
  least-squares solver will perform.

  In this paper we propose an alternative error function obtained by removing 
  some non-linearities from the standard used one - i.e. the geodesic error 
  function. 
  Comparative experiments conducted on common benchmarking datasets confirm 
  that our function is more robust to noise that affects the rotational 
  component of the pose measurements and, thus, exhibits a larger convergence 
  basin than the geodesic. 
  Furthermore, its implementation is relatively easy compared to the geodesic 
  distance. This property leads to rather simple derivatives and nice numerical 
  properties of the Jacobians resulting from the effective 
  computation of the quadratic approximation used by Gauss-Newton algorithm.
\end{abstract}

\section{Introduction}\label{sec:intro}
Simultaneous Localization and Mapping (SLAM) is a well known problem
that has been studied intensively by the research community over  the
last two decades. Many paradigms have been proposed through the years to
efficiently solve this problem. Amongst them, the \textit{graph-based} approach 
gained much popularity in the last decade thanks to its efficiency and 
flexibility.

Graph-based SLAM approaches have generally two main components:
a front-end whose role is to construct an abstract pose-graph from raw 
measurement data, and a back-end that has the task to provide the front-end and 
potentially other modules with an up-to-date consistent configuration of the 
pose-graph. For a detailed overview on this paradigm, we refer the reader to 
the work of Grisetti~\emph{et al.}~\cite{grisetti2010tutorial}.

A pose-graph is a representation of a stochastic map. Its nodes represent
samples of the robot trajectory or locations of local maps. Edges represent
spatial constraints between local maps that can be inferred from measurements.
As the robot travels, the graph is augmented by adding new nodes and edges,
and its configuration might become inconsistent.

The task of the back-end is to constantly provide a consistent
configuration of the pose-graph. The problem of graph optimization has
been deeply investigated by the community in the recent years and
effective systems are available.  Nowadays, state-of-the-art back-ends
are iterative solvers based on least-squares optimization \cite{ceres-solver} 
\cite{kummerle2011g} \cite{dellaert2012gtsam}.  These
solvers require that the current estimate is reasonably close to the
optimum, assumption generally verified while running SLAM
incrementally.  Least-squares solvers operate by iteratively solving a
quadratic approximation of the original minimization problem. A better
approximation results in both a larger convergence basin and in
faster convergence.  
However, the objective function for pose-to-pose edges in
a pose graph is highly non-linear, especially in the 3D case, due to the presence of 
rotations.

\begin{figure}[!t]
  \centering
  \begin{subfigure}{0.7\columnwidth}
    \includegraphics[width=0.9\linewidth]{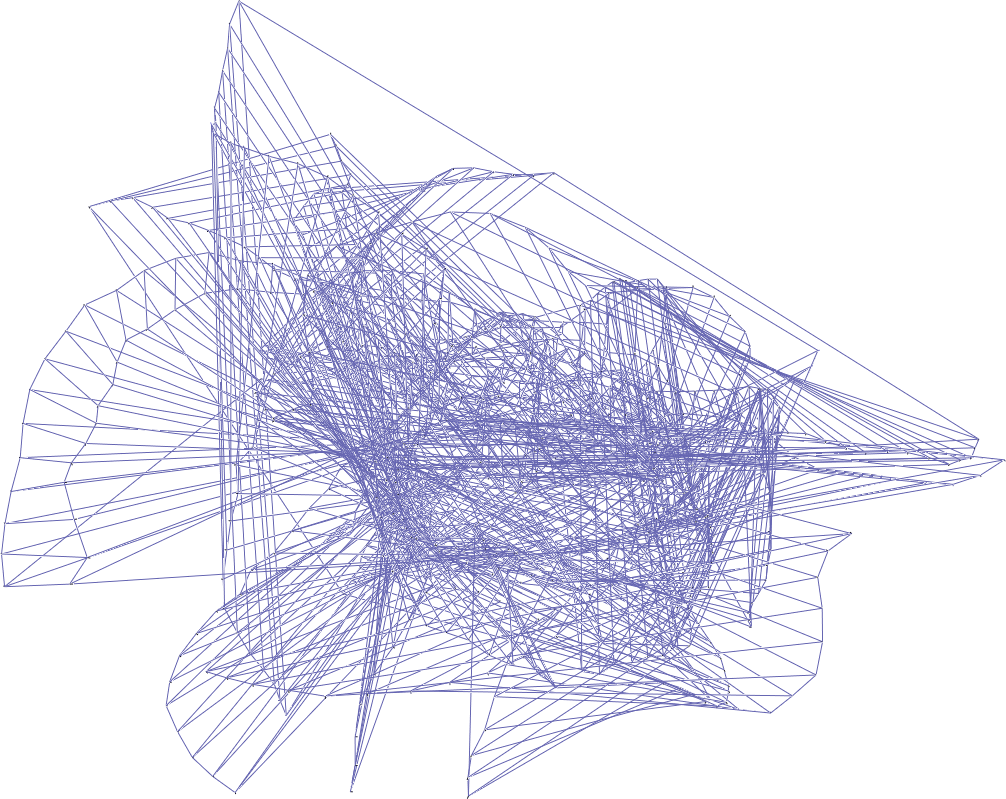}
    \subcaption{Initial guess from the spanning tree.}
    \label{fig:sim-torus3d-all-rot-spanning}
  \end{subfigure} \\ \vspace{15pt}
  \begin{subfigure}{0.45\columnwidth}
    \includegraphics[width=0.9\linewidth]{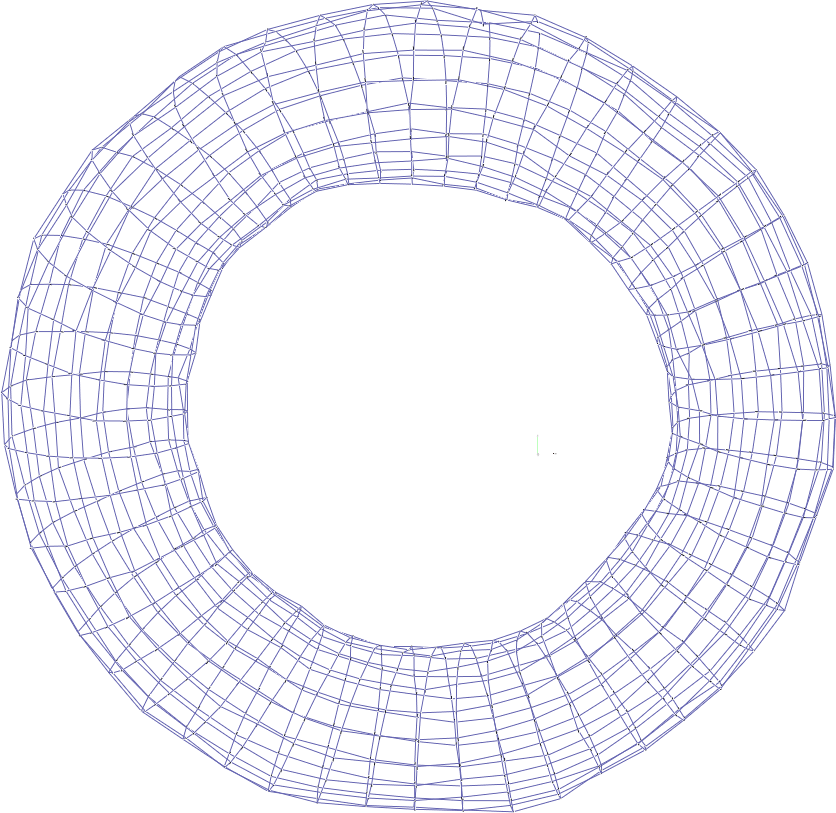}
    \subcaption{Our approach.}
    \label{fig:sim-torus3d-all-rot-spanning-out-lm-chord}
  \end{subfigure}  \hspace{5px}
  \begin{subfigure}{0.45\columnwidth}
    \includegraphics[width=0.9\linewidth]{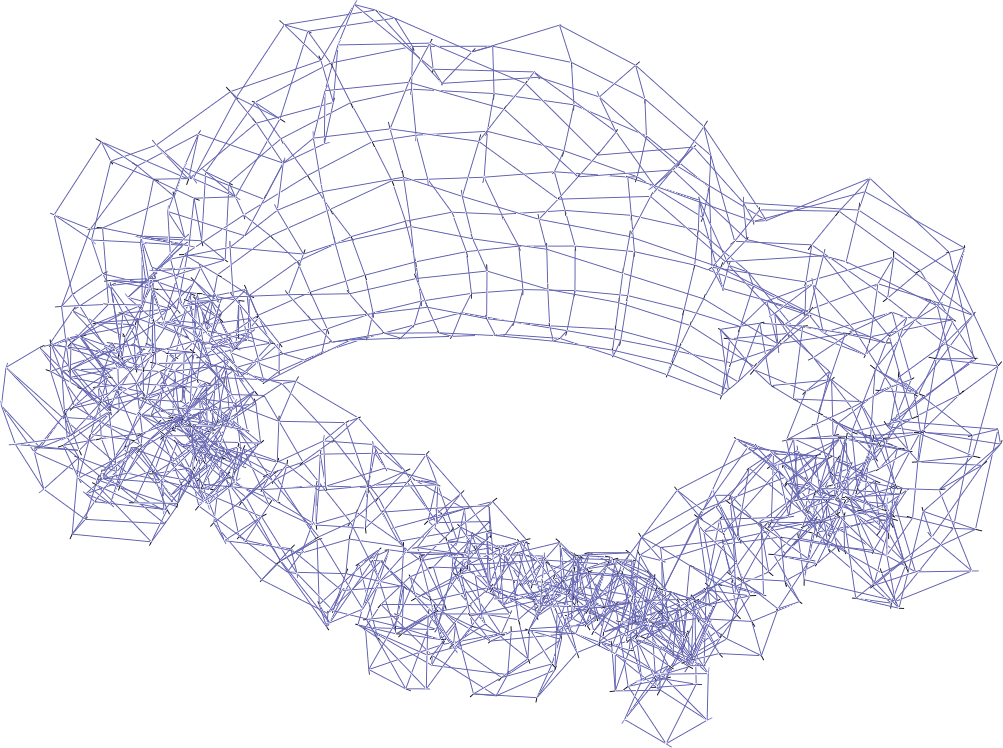}
    \subcaption{Geodesic distance.}
    \label{fig:sim-torus3d-all-rot-spanning-out-lm-geo}
  \end{subfigure}
  \caption{Result of optimization on the \texttt{torus-b} using 
  Levenberg-Marquardt algorithm - 100 iterations. We perturbed the dataset with 
  noise $\Sigma_R = [0.1\;0.1\;0.1]$ and $\Sigma_t = [0.001\;0.001\;0.001]$ - 
  added respectively on the rotational and translational part of the pose. 
  Optimization using the geodesic distance leads to a local minimum, while our 
  approach succeeds in finding the right nodes configuration.}
  \label{fig:motivational}
  \vspace{-20pt}
\end{figure}

To deal with this problem, Carlone~\emph{et al.}~\cite{Carlone11rss-lagoPGO2D} 
proposed an approach to for solving 2D pose-graphs that constructs an exact
quadratic approximation of the original problem by means of
``unwinding'' the angular component of the pose differences to avoid
singularities. Such a solution is reported to provide good results when dealing 
with spherical covariances.

However such approach cannot be easily adapted to the three
dimensional case.  To this extent in a recent work Carlone~\emph{et
  al.}~\cite{carlone2015initialization} addressed the crucial issue of
finding a good initial guess for 3D pose-graph optimization.
Determining a good initial guess is crucial when the system is started
from an unknown configuration of the poses - e.g. when
operating on unorganized data.
However, in case of on-line SLAM a reasonable solution is typically
available. Still, in all cases the problem is turned to a
non-linear optimization.  A pose-graph is a particular case of a
factor graph where the edges represent binary factors.

Several publicly available tools are commonly used to solve factor
graphs such as GT-SAM~\cite{dellaert2012gtsam}, 
Ceres-Solver~\cite{ceres-solver} and \g2o~\cite{kummerle2011g}.
To use these tools one needs just to describe the domain of the state
variables (nodes) and how to compute the errors induced by the
measurements (factors) used in the problem.  The additive nature of
the factors allows the user to compose multiple heterogeneous
measurements in a single factor graph.

In this work we propose a different error function to model
pose-to-pose measurements that exhibits a smoother behavior compared to
the commonly used geodesic distance. Its main features are:
\begin{itemize}
  \item[--] enlarged basin of convergence and, as a consequence, increased 
  robustness of the optimization process
  \item[--] simpler derivatives - compared the geodesic distance - that can be 
  easily computed in closed form.
\end{itemize}
As illustrated in~\figref{fig:motivational}, our approach succeeds in finding 
the optimal nodes configuration in cases where the geodesic function remains 
stuck in local minimum. 
Moreover, when measurements are affected by realistic noise, the optimum 
obtained with our approach is equivalent to the one retrieved with the geodesic 
distance. 
Our claims are supported by comparative experiments on publicly available 
datasets. 
In addition to that, we provide an open source plugin\footnote{\scriptsize 
Source code: \url{https://srrg.gitlab.io/g2o_chordal_plugin.html}} for \g2o 
that implements our error function, allowing to reproduce all the proposed 
experiments. In addition to that, we provide an Octave implementation of a 
simple least squares system, which is used to teach a SLAM course at Sapienza 
University of Rome.

\section{Related Work}\label{sec:related}
The work of Lu and Milios~\cite{lu1997globally} defined the
graph-SLAM techniques in the context of laser scans.  In this work
they construct a pose-graph where each node represents a robot
pose and the laser scan acquired at that pose.  Edges were obtained
either by odometry or by registering scans that were acquired at
nearby poses. To optimize such a pose-graph they employed Gauss-Newton
and treated the 2D poses as 3D Euclidean vectors, handling the
singularities arising from angular difference in an ad-hoc manner. In
the solution of the linear problem within least squares, the authors
disregarded the sparse nature of the resulting linear system.  This
was not seen as an issue, since the number of scans considered and,
thus, the size of the linear system was rather small, however as the
size of the problem increases the solution of the linear system
quickly become a computational bottleneck.

Gutmann and Konolidge~\cite{gutmann1999incremental} addressed the problem of
incrementally building a map, finding topological relations and loop
closures based on local maps, and triggering the optimization
only when the current state of the graph becomes substantially inconsistent.
To avoid unnecessary computation they restricted the optimization to
the sole portion of the graph that was reported as inconsistent,
thus trading off computation and optimality of the solution.

To approach the computational issues in least-squares optimization
Howard~\emph{et. al}~\cite{Howard2001relaxation} and Duckett~\emph{et 
al.}~\cite{Duckett02Mapping} introduced relaxation. 
This approach is reported to be easy to implement, however its convergence rate 
is linear instead of quadratic. 
Compared to least squares approaches, each iteration is
faster but more iterations are required to find the optimum.
Frese~\emph{et al.} proposed to use multilevel 
relaxation~\cite{Frese04relaxation} to increase the convergence speed
of the method.

Olson \emph{et al.}~\cite{olson2006fast}, proposed to
use \textit{Stochastic Gradient Descent} instead of least squares for 2D environments.
Subsequently, this work was extended by Grisetti \emph{et 
al.}~\cite{grisetti2007tree} addressing the 3D case and the introduction of a 
tree-based parameterization for the problem that further increased the
convergence speed. However this work assumes that the measurements 
covariances are spherical and, therefore, this approach is not general.

Dallaert \emph{et al.}~\cite{dellaert2006square} released a system 
known as $\sqrt{SAM}$ that exploited the sparsity of the linear system to efficiently
compute a solution.  In the same line, Kaess \emph{et al.} proposed
iSAM~\cite{kaess2007isam} and iSAM2~\cite{kaess2012isam2}. These two
works leverage on $\sqrt{SAM}$, adding respectively the features of
incremental optimization and new data structures to the original
system configuration.  In parallel K\"{u}mmerle \emph{et al.} proposed 
\g2o~\cite{kummerle2011g}, an optimization tool designed to easily 
prototype sparse least-squares solvers for factor graphs. 
\g2o builds on concepts from operating system realizing a layered
structure that separates the problem definition from the problem
solution and implements a plugin architecture that allows to modify most of its 
components. This allows the user to apply heterogeneous strategies to
solve the factor graph, and to extend the types of ``factors'' and
``node variables'' upon need.

To further address the issues of poor initial guess and scalability,
Ni~\emph{et. al}~\cite{NiICRA2007} and subsequently Grisetti~\emph{et
  al}~\cite{grisetti12iros} applied divide and conquer strategies to
find the optimal solution. The first approach leverages on nested
dissection to solve the linear system, while the latter assembles a
set of non-linear sparser problems from local portions of the graph.

When used to solve pose-graphs all those approaches suffer
from the non-linearities introduced by the rotational component of the problem, 
leading to weak convergence results when the initial guess has a noisy rotational part. 
Notabily, Carlone \emph{et al.} investigated this
issue~\cite{carlone2015initialization}, proposing to relax the
rotational constraints using different distance, generating a better
initial guess for the standard optimization.

In this paper we propose an error function for 
pose-to-pose constraints that improves the stability of the optimization 
process that can be used in arbitrary pose-graphs.
The proposed function is relatively easy to implement and has nice numerical 
properties.  
Our contribution is orthogonal to all least-squares methods mentioned above and 
can be used in conjunction with them.

\section{Pose Graphs Optimization}\label{sec:pose-graphs}
In this section we quickly review some concepts on non-linear
optimization for pose graphs.  To deal with the non-Euclidean objects
such as the isometries in the factor graph we rely on the manifold
encapsulation technique proposed by
Hertzberg~\emph{et. al}~\cite{hertzberg2012tutorial}.  We furthermore
discuss the effect of non-euclidean domains when evaluating the error
function.

As stated in the introduction, the pose-graph is a graph whose nodes represent
robot poses and edges represent relative transformations between poses.
Let $\bX=\bX_{1:N}$ be the nodes in the graph, represented as 2D or 3D 
isometries, and let $\{ \left<\bZ_{ij}, \bOmega_{ij} \right> \}$ be the edges 
in the graph with the subscript indicating the connected nodes.
To capture the stochastic nature of the measurement to an edge we
store not only the isometry $\bZ_{ij}$ that represents the measured relative 
location between nodes $i$ and $j$, but also an information matrix 
$\bOmega_{ij}$ that captures the measurement's uncertainty along the different 
dimensions.

Pose-graph optimization consists in finding the configuration $\bX^*$ of nodes that minimizes
the following objective function
\begin{equation}
  \bX^* = \argmin_{\bx} \sum_{i,j} \| \be_{ij}(\bX_i,\bX_j) \|_{\bOmega_{ij}}
  \label{eq:objective-fn}
\end{equation}
Here $\be_{ij}(\bX_i, \bX_j)$ is a vector function that measures the difference between the predicted measurement
$\hat \bZ_{ij}=\bh(\bX_i, \bX_j)=\bX_j \ominus \bX_i$ and the measurement $\bZ_{ij}$. With $\ominus$ we refer to the motion
decomposition operator as introduced in~\cite{Smi90EstABBREV}.
Assuming all variables are vectors, a straightforward implementation of the error function is thus the following:
\begin{equation}
  \be_{ij}(\breve \bX)=\bh(\breve \bX_i, \breve \bX_j) - \bZ_{ij}. \label{eq:error-euclidean}\\
\end{equation}
\eqref{eq:objective-fn} is usually solved by iterative
non-linear least squares minimization, leading to the popular Gauss-Newton
or Levemberg-Marquardt methods.
We refer the reader to~\cite{grisetti2010tutorial} for a comprehensive tutorial 
on on least-squares on pose-graphs.
The core idea of these methods is to repeatedly refine a current initial guess 
of the solution $\breve \bX$ by solving many times its quadratic approximation.
The latter is obtained through the first-order Taylor expansion of the error function evaluated around $\breve \bX$:

\begin{align}
  \be_{ij}(\breve \bX + \bDeltaX) &= \bh(\breve \bX_i+\bDeltaX_i, \breve \bX_j+\bDeltaX_j) - \bZ_{ij} \label{eq:error-euclidean-pert}\\
  &\approx \be_{ij}(\breve \bX)
             + \left.\frac{\partial \bh_{ij}(\bX_i, \breve \bX_j)}{\partial \bX_i}\right|_{\bX_i=\breve\bX_i}\bDeltaX_i + \nonumber \\
             &+ \left.\frac{\partial \bh_{ij}(\breve \bX_i, \bX_j)}{\bX_j}\right|_{\bX_j=\breve\bX_j}\bDeltaX_j \nonumber
\end{align}

\subsection{Smooth Manifolds Encapsulation}
The above operation leverages on the correct definition of vector subtraction and addition,
and assumes that both states $\bX$ and measurements $\bZ$ live in Euclidean spaces.
In case of pose graphs, however this is no longer the case since isometries lie on smooth
manifolds~$SE(2)$ and~$SE(3)$ respectively. A manifold is a space
that, albeit non homeomorphic to~$\mathbb{R}^n$, admits a locally
Euclidean parametrization around each element~$\bM$ of the domain,
commonly referred to as \textit{chart}. Therefore, a chart computed around
a manifold point $\bM$ is a function from $\mathbb{R}^n$ to a new
point $\bM'$ on the manifold:
\begin{equation}
  \chart_\bM(\bDeltam):\mathbb{R}^n \rightarrow \mathbb{M}. 
  \label{eq:chart}
\end{equation}
Intuitively, $\bM'$ is obtained by ``walking'' along the
perturbation $\bDeltam$ on the chart, starting from the chart origin. A null
motion ($\bDeltam=\mathbf{0}$) on the chart, leaves us at the point where the
chart is constructed:~$\chart_\bM(\mathbf{0})=\bM$.

Similarly, given two points $\bM$ and $\bM'$ on the manifold, we can
determine the motion $\bDeltam$ on the chart constructed around $\bM$
that would bring us to $\bM'$. Let this operation be the inverse
$\chart^{-1}_\bM(\bM')$.  The direct and inverse charts allow us to
define operators on the manifold that are analogous to the sum and
subtraction in the Euclidean space. Let $\boxplus$ and~$\boxminus$ be
those operators, defined as follows:
\begin{align}
  \label{eq:generic-box-plus}
  \bM \boxplus \bDeltam &\triangleq \chart_\bM(\bDeltam)\\
  \label{eq:generic-box-minus}
  \bM' \boxminus \bM &\triangleq \chart^{-1}_\bM(\bM')
\end{align}
This notation was first introduced by Hertzberg and
Frese~\cite{hertzberg2012tutorial}, and allows us to easily adapt the
Euclidean version of non-linear Least-Squares to operate on manifold
spaces. The parameterization of the chart is usually chosen to be of
minimal dimension, while the representation of the manifold
element~$\mathbf{M}$ can be chosen arbitrarily.  Accordingly, two
possible parametrizations for SE(3) objects are:
\begin{align}
  \label{eq:extended-parametrization}
  \bX = \begin{pmatrix} \bR & \bt \\ \bZero_{3 \times 1} & 1\end{pmatrix} \quad \bR &= \bR_x(\phi)\,\bR_x(\theta)\,\bR_x(\psi) \\
  \label{eq:minimal-parametrization}
  \bDeltax &= \begin{pmatrix}x&y&z&\phi&\theta&\psi\end{pmatrix}^{T}
\end{align}

Accordingly, to compute the difference
between two isometries or to apply an increment to an isometry, we need to
define the operators~$\boxminus$ and~$\boxplus$. In the remainder of
this section, we will use the following definition for such operators:
\begin{align}
  \label{eq:boxplus-v2t}
  \bX \boxplus \bDeltax &= \text{v2t}(\bDeltax)\bX \\
  \label{eq:boxminus-t2v}
  \bX_a \boxminus \bX_b &= \text{t2v}(\bX_b^{-1}\bX_a)
\end{align}
Here~$\text{t2v}$ and~$\text{v2t}$ map an isometry into a
6D minimal vector and vice-versa. We refer the reader to
Appendix~\ref{sec:appendix} for the mathematical definitions of these
functions. Hence, we can compute the error between predicted and
actual measurement as~$\be_{ij} = \hat\bZ_{ij} \boxminus
\bZ_{ij}$. To minimize the objective function in~\eqref{eq:objective-fn} using 
an iterative approach we need
to compute its Taylor approximation around the current estimate $\breve \bX$.
Setting $\be_{ij}=\be_{ij}(\breve \bX)$ and expressing the perturbation on the 
charts results in the following expansion:
\begin{align}
  \label{eq:manifold-error}
  \be_{ij}(\breve \bX &\boxplus\bDeltax) = \bh_{ij}(\breve \bX_i\boxplus\bDeltax_i, \breve \bX_j\boxplus\bDeltax_j) \boxminus \bZ_{ij} \\
  &\approx \be_{ij} +\underbracket{\frac{\partial \be_{ij}(\breve \bX_i\boxplus\bDeltax_i, \breve \bX_j)}{\partial\bDeltax_i}\bigg\rvert_{\bDeltax_i=0}}_{\tilde \bJ_{i}} \bDeltax_i +   \label{eq:jac-i-increments} \\
  &+\underbracket{\frac{\partial \be_{ij}(\breve \bX_i, \breve \bX_j\boxplus\bDeltax_j)}{\partial\bDeltax_j}\bigg\rvert_{\bDeltax_j=0}}_{\tilde \bJ_{j}} \bDeltax_j 
  \label{eq:jac-j-increments}
\end{align}
The smoother the function $\be_{ij}(\cdot)$ with respect to the
perturbation, the better the final quadratic form will approximate
the nonlinear problem. This results both in less iterations and larger
convergence basin. To the limit, if the Jacobians are not affected by
the linearization point one can find the solution in just one
iteration.
In~\eqref{eq:manifold-error} we explicitly addressed the fact that only the 
blocks $\bDeltax_i$ and $\bDeltax_j$ in the perturbation vector 
$\bDeltax=\bDeltax_{1:N}$ determine
the error between nodes $i$ and $j$. The full jacobian with respect to all 
perturbation blocks has the following general structure:
\begin{equation}
  \tilde \bJ_{ij} = \left[\bZero \cdots \bZero\, \bJ_i \, \bZero \cdots \bZero 
  \, \bJ_j \, \bZero\cdots \bZero\right].
\end{equation}

\subsection{Error on a Chart}
Comparing equations~\eqref{eq:error-euclidean} and~\eqref{eq:manifold-error}, the reader might notice that the subtraction between prediction $\hat \bZ_{ij}=\bh(\bX_i,\bX_j)$ and observation $\bZ_{ij}$ has been
replaced  by a $\boxminus$ operator. This is coherent with the fact that the 
measurement $\bZ_{ij}$ and the prediction $\bh_{ij}$ are manifolds.
This, however, introduces an additional nonlinear transformation in the 
calculation of the omega-norm. Intuitively, since the error is computed on a 
chart constructed around the measurement,
the value of the error on the chart needs to be reestimated each time the prediction changes.
This is consistent with the fact that the original information matrix of the measurement $\bOmega_{ij}$ has dimensions consistent with the measurement $\bZ_{ij}$, which might be different
from the ones of the error vector $\be_{ij}$. This can be solved by computing a 
Gaussian approximation of the error distribution around the manifold 
measurement: given the relations $\bz = \text{t2v}(\bZ)$ and $\bz_{ij} = 
\text{t2v}(\bZ_{ij})$ we can write:
\begin{align}
  p(\bz) &\sim \cN(\bz_{ij}, \; \bOmega_{ij}^{-1})\\
  \be_{ij}&= \hat \bZ_{ij} \boxminus \bZ\\ 
  p(\be_{ij}) &\sim \cN(\hat \bZ_{ij} \boxminus \bZ_{ij},\; \bJ_{\bZ_{ij}}\bOmega_{ij}^{-1}\bJ_{\bZ_{ij}}^T) \label{eq:error-distribution}\\
  \text{with} \quad \bJ_{\bZ_{ij}} &= \left.\frac{\partial (\hat \bZ_{ij} 
  \boxminus \bZ)}{\partial \bz} \right|_{\bz = {\bz_{ij}}} 
  \label{eq:jac-measurements}
\end{align}
The reader might notice that $\bJ_{\bZ_{ij}}$ depends on the
prediction and, thus, on the linearization point. Accordingly, the
covariance of the error $\tilde \bOmega_{ij}^{-1} =
\bJ_{\bZ_{ij}}\bOmega_{ij}^{-1}\bJ_{\bZ_{ij}}^T$ is a function of the
state and should be recomputed at each iteration.  However, when using
the same representation for the error vector and the perturbations,
and when the prediction and the measurement are close we have that the
Jacobian $\bJ_{\bZ_{ij}} \approxeq \bI$ and many state-of-the-art
systems simply ignore this step.

\subsection{Gauss-Newton for Pose Graphs on a Manifold}
For sake of completeness, in this section we report an algorithmic
presentation of the minimization algorithm that combines all the elements
sketched in the previous sections.
\algref{alg:gn-manifolds} reports the pseudo-code of such optimization process.

\begin{algorithm}[!t]
  \caption{Gauss-Newton minimization algorithm for manifold
    measurements and state spaces}
  \begin{algorithmic}[1]
    \Require{
      Initial guess $\bXlin$;
      Measurements $\mathcal{C} =   \{\langle\bZ_k, \bOmega_k\rangle\}$
    }
    \Ensure{Optimal solution $\bX^\star$}
    \State $F_{new} \leftarrow \breve{F}$ \Comment{compute the current error}
    \While{$\breve{F} - F_{new} > \epsilon$}
    \State $\breve{F} \leftarrow F_{new}$
    \State $\mathbf{b} \leftarrow 0$
    \State $\bH \leftarrow 0$
    \For{$\bZ_{ij} \in \mathcal{C}$}
    \State $\hat \bZ_{ij} \leftarrow \bh_{ij}(\bXlin)$ \Comment{compute prediction}
    \State $\be_{ij} \leftarrow  \hat \bZ_{ij} \boxminus \bZ_{ij}$ \Comment{compute the error}
    \State $\tilde{\bJ}_{ij} \leftarrow \frac{\partial \tilde{\be}_k
      (\bh_k(\bX \boxplus \bDeltax),\bz_k)}{\partial \bDeltax_k} \big\rvert_{\bDeltax_k = 0}$ \Comment{jac of~$\boxplus$}
    \State $\bJ_{\bZ_{ij}} \left.\frac{\partial (\hat \bZ_{ij} \boxminus \bZ)}{\partial \bZ} \right|_{\bZ = {\bZ_{ij}}}$ \Comment{error jac. on the chart}
    \State $\tilde{\bOmega}_k \leftarrow \left(\bJ_{\bZ_k}\bOmega_k\bJ_{\bZ_k}^{T}\right)^{-1}$ \Comment{remap Omega}
    \State $\bH_k \leftarrow \bJ_k^T\tilde{\bOmega}_k\bJ_k$ \Comment{contribution of~$\bZ_{ij}$ in~$\bH$}
    \State $\mathbf{b}_k \leftarrow \bJ_k^T\tilde{\bOmega}_k\be_k$ \Comment{contribution of~$\bZ_{ij}$ in~$\bb$}
    \State $\bH \pluseq \bH_k$ \Comment{accumulate contributions}
    \State $\mathbf{b} \pluseq \mathbf{b}_k$ \Comment{accumulate contributions}
    \EndFor
    \State $\bDeltax \leftarrow solve(\bH\bDeltax = -\mathbf{b})$ \Comment{solve w.r.t.~$\bDeltax$}
    \State $\bXlin \leftarrow \bXlin \boxplus \bDeltax $ \Comment{update the state}
    \State $F_{new} \leftarrow F(\bXlin)$ \Comment{compute the new error}
    \EndWhile
    \State
    \Return $\bXlin$
  \end{algorithmic}
  \label{alg:gn-manifolds}
\end{algorithm}
The quadratic form is obtained by expanding the Taylor approximation in the 
summands of~\eqref{eq:objective-fn} as follows:

\begin{align}
  {}&\mathbf{F}_{ij}(\breve\bX \boxplus \bDeltax) = \|\be_{ij}(\bX \boxplus \bDeltax)\|_{\tilde \bOmega_{ij}} \nonumber \\ 
  &\approx (\be_{ij} + \bJ_{ij}\bDeltax)^{T} \tilde\bOmega_{ij} (\be_{ij} + \bJ_{ij}\bDeltax) \nonumber =\\
  &=\bDeltax^T\underbracket{\tilde\bJ_{ij}^T\tilde\bOmega_{ij}\tilde\bJ_{ij}}_{\bH_{ij}}\bDeltax + 2\,\underbracket{\be_{ij}\tilde\bOmega_{ij}\tilde{\bJ_{ij}}}_{\bb_{ij}}\bDeltax + \underbracket{\be_{ij}^T\tilde\bOmega_{ij}\be_{ij}}_{\bc_{ij}}
  \label{eq:cost-ij}
\end{align}

Considering all the measurements, the global cost around $\breve \bX$ as a function of the perturbation function will be:
\begin{align}
  \mathbf{F}(\breve{\bX} &\boxplus \bDeltax) = \sum_{\bZ_{ij}\in\mathcal{C}}\mathbf{F}_{ij}(\breve\bX \boxplus \bDeltax) \nonumber \\ 
  &\approx \bDeltax^T\bH\bDeltax + 2\,\bb\bDeltax + \bc
  \label{eq:cost-function-lin}
\end{align}
We can find the minimum of~\eqref{eq:cost-function-lin} computing its derivative and equating it to 0. This means that we have to solve the following linear system w.r.t.~$\bDeltax$
\begin{equation}
  \bH \bDeltax = -\bb
  \label{eq:linear-sys}
\end{equation}
The result will be an increment~$\bDeltax$ that applied to~$\breve \bX$ will 
lead to a state closer to the optimal one:
\begin{equation}
  \bX \leftarrow \breve \bX \boxplus \bDeltax
  \label{eq:increment}
\end{equation}
Iterative algorithms repeat this process until convergence is reached.

\section{Pose Error Functions}\label{subsec:error-functions}
In this section, we analyze in depth a typical error function used in pose-graph optimization, and we will focus on the 3D case.
The extension to 2D pose graphs is straightforward.

\subsection{Standard SE3 Error}\label{par:standard}
A standard way of computing the pose-pose error uses the operator~$\boxplus$ and~$\boxminus$ defined in~\eqref{eq:boxplus-v2t} and~\eqref{eq:boxminus-t2v}. Following this formalization and embedding the perturbations together with the~$\boxplus$ operator, we can compute the perturbed error as:
\begin{align}
  \be_{ij}&(\bX_i\boxplus\bDeltax_i, \bX_j\boxplus\bDeltax_j) = \nonumber \\
  {} &= \text{t2v}\left(\bZ_{ij}^{-1}\left(\text{v2t}(\bDeltax_i)\,\bX_i\right)^{-1} \,
  \left(\text{v2t}(\bDeltax_j)\,\bX_j\right)\right)
  \label{eq:t2v-perturbed-error}
\end{align}
\eqref{eq:t2v-perturbed-error} is highly non-linear and it suffers form a large 
number of singularities, mainly due to the use of function t2v that converts a 
transformation matrix in a minimal representation - refer to 
Appendix~\ref{sec:appendix}. 
Therefore, it propagates such non-linearities in the Jacobians
and, thus, to the whole optimization process.

\subsection{Chordal-Based SE3 Error}\label{par:chordal}
As mentioned in~\cite{carlone2015initialization}, we can define an alternative 
error function based on the concept of \textit{chordal distance}. To this end, 
we first introduce the function $\text{flatten}(\cdot)$, defined as follows:
\begin{equation}
  \text{flatten}(\bT) = \begin{pmatrix}
  \br_1^T&\br_2^T&\br_3^T&\bt^T\end{pmatrix}^T
  \label{eq:flatten}
\end{equation}
where~$\br_j$ represents the $j$-th versor of the rotation matrix~$\bR$. 
Basically, it is a linear transformation that reshapes an isometry into the 
12D-vector containing rotation vector and translation. 
According to this, we define new~$\boxplus$ and~$\boxminus$ operators as follows:
\begin{align}
\bX \boxplus \bDeltax &= \text{v2t}(\bDeltax)\bX \\ \bX_a \boxminus \bX_b &=
\text{flatten}(\bX_a) - \text{flatten}(\bX_b)
\label{eq:boxminus-chordal}
\end{align}
It is important to notice that in~\eqref{eq:boxminus-chordal}, the difference is done through the
standard Euclidean minus operator. As a result, the 12-dimensional error between two $SE(3)$ becomes:
\begin{align}
  \be_{ij} = \bZhat_{ij} \boxminus \bZ_{ij} = \text{flatten}(\bX_i^{-1}\bX_j)
  - \text{flatten}(\bZ_{ij})
  \label{eq:error-chordal}
\end{align}
Note that, in this sense, \emph{we use two different parametrizations
  for the error and increments}. The derivation of Jacobians $\bJ_i$
and $\bJ_j$ from~\eqref{eq:error-chordal} is reported in
Appendix~\ref{sec:appendix}. Eliminating $\text{t2v}$ 
from~\eqref{eq:t2v-perturbed-error} removes substantial non-linearities
and, thus, produces a smoother function. 
This leads to an enlarged convergence basin, increasing the robustness of the 
optimization process with respect to noise.  Finally, we observe that the two
Jacobians are linked by the relation $\bJ_i = -\bJ_j$, as reported in 
Appendix~\ref{sec:appendix}.  Accordingly,
the four contributions to~$\bH$ introduced by measurement~$\bZ_{ij}$
will be:
\begin{equation}
  \bH_{ii} = \bH_{jj} = -\bH_{ij} = -\bH_{ji} = \bJ_i^T\bOmega_{ij}\bJ \label{eq:jac-opposite}
\end{equation}
The consequences of~\eqref{eq:jac-opposite} are rather substantial
in the computation of the $\bH$ matrix. In one single operation we can
compute all four entries of $\bH$ that are affected by a measurement,
and this leverages the cost of operating with twelve instead of six
dimensional error vectors.
\begin{figure}[!t]
    \centering
    \includegraphics[width=0.99\columnwidth]{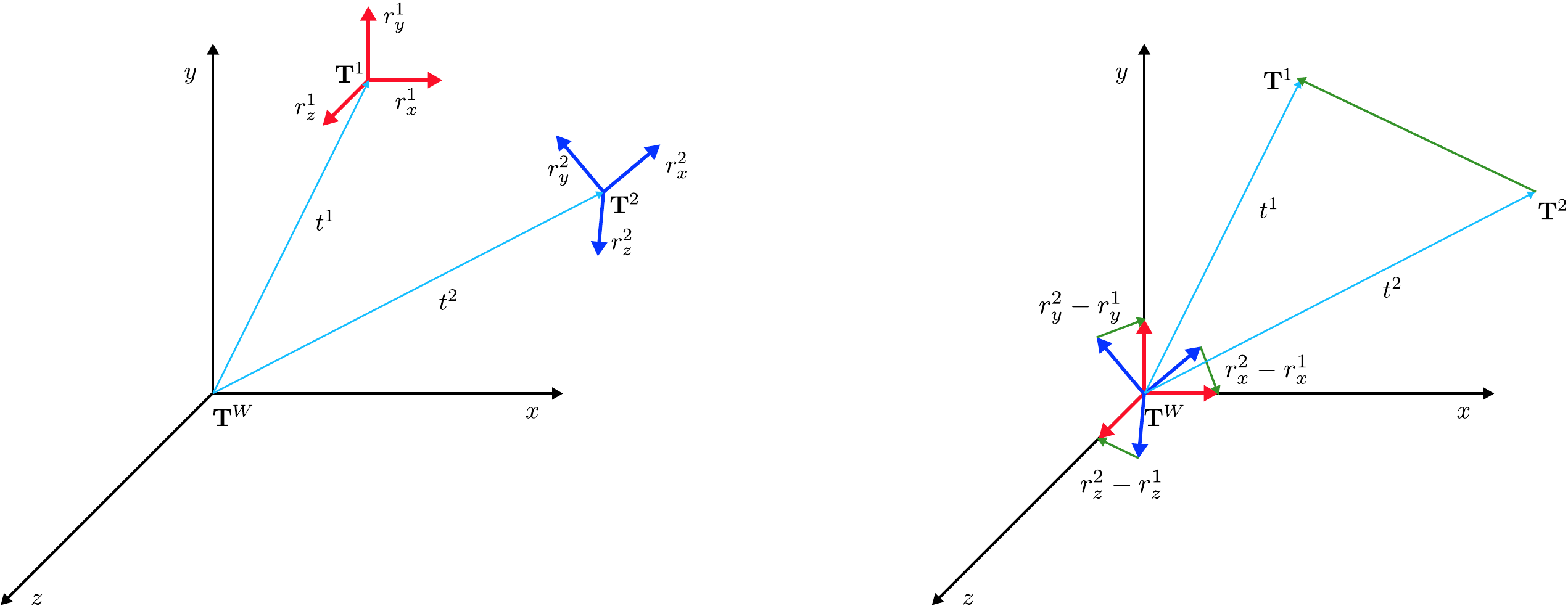}
    \caption{In this figure we show the principle behind our \textit{SE(3)} error function. The leftmost illustration depicts two poses $\bT_1$ and $\bT_2$. Supposing that we want to compute~$d_{chord}(\bT_1, \bT_2)$, the rightmost illustration visually shows how this is computed: the translational part is simply $\bt_2 - \bt_1$ as usual; the rotational part is computed as the difference between the versors of the two rotations - namely $\br_j^2 - \br_j^1$ with $j = \{x,y,z\}$}
    \label{fig:chordal-distance}
    \vspace{-15pt}
\end{figure}
Notably,~\eqref{eq:error-chordal} uses a vector difference instead of the 
non-linear t2v, therefore the information matrix $\tilde \bOmega_{ij}$ does not 
need to be recomputed at each iteration. This has the dual effect of speeding 
up the computation and leading to a more consistent quadratic approximation of 
the problem.

Usually, the input problem expresses the measurements trough a minimal 
six-dimensional parameterization such as translation and normalized quaternion
or translation and Euler angles.
Therefore, we cannot reuse these information matrices as they are, but we
need to transform them to the new representation that has 12 parameters.
This can be done using either first order error propagation
or with the Unscented Transform.
Mapping a 6 dimensional Gaussian onto a 12 dimensional space
will unavoidably lead to a non positive definite covariance matrix
due to the inherent rank losses.
We solve this problem adding a small $\epsilon > 0$ to the null 
singular values of the covariance matrix before inverting it to obtain the 12D 
information matrix.
We verified this procedure by performing the inverse transformation (12D to 6D)
and by verifying that the restored problem has information matrices
numerically close to those of the original one.

\section{Experimental Evaluation}\label{sec:exp}
In this section we investigate the effects the \textit{chordal distance} error 
function presented in Section~\ref{par:chordal}.  
We provide some key tests to support the claim
that our error function leads to a larger convergence basin with respect to
the one based on the geodesic distance.  To this end, we tested the
optimization on several standard pose-graph datasets, comparing the
evolution of the optimization residual error employing the standard
and the proposed error function. All tests have been conducted on
3-dimensional pose graphs.

\begin{table}[b]
  \centering
  \begin{tabular}{|c|c|c|}
    \hline
    \rowcolor{SeaGreen3!30!} \textbf{Dataset}  & \textbf{\# Vert.} & \textbf{\# 
    Meas.} \\ [0.5ex] \hline \hline
    \texttt{garage} & 1661 & 6275 \\ \hline
    \texttt{grid} & 8000 & 22236 \\ \hline
    \texttt{sphere-a} & 2200 & 8647 \\ \hline
    \texttt{torus-a} & 5000 & 9048 \\ \hline
    \texttt{sim-manhattan} & 5001& 60946 \\ \hline
    \texttt{sphere-b} & 2500 & 9799 \\ \hline
    \texttt{torus-b} & 1000 & 1999 \\ \hline
  \end{tabular}
  \caption{Specs of the datasets used for the experiments.}
  \label{tab:dataset-specs}
\end{table}

\begin{figure}[!t]
  \centering
  \includegraphics[width=0.9\columnwidth]{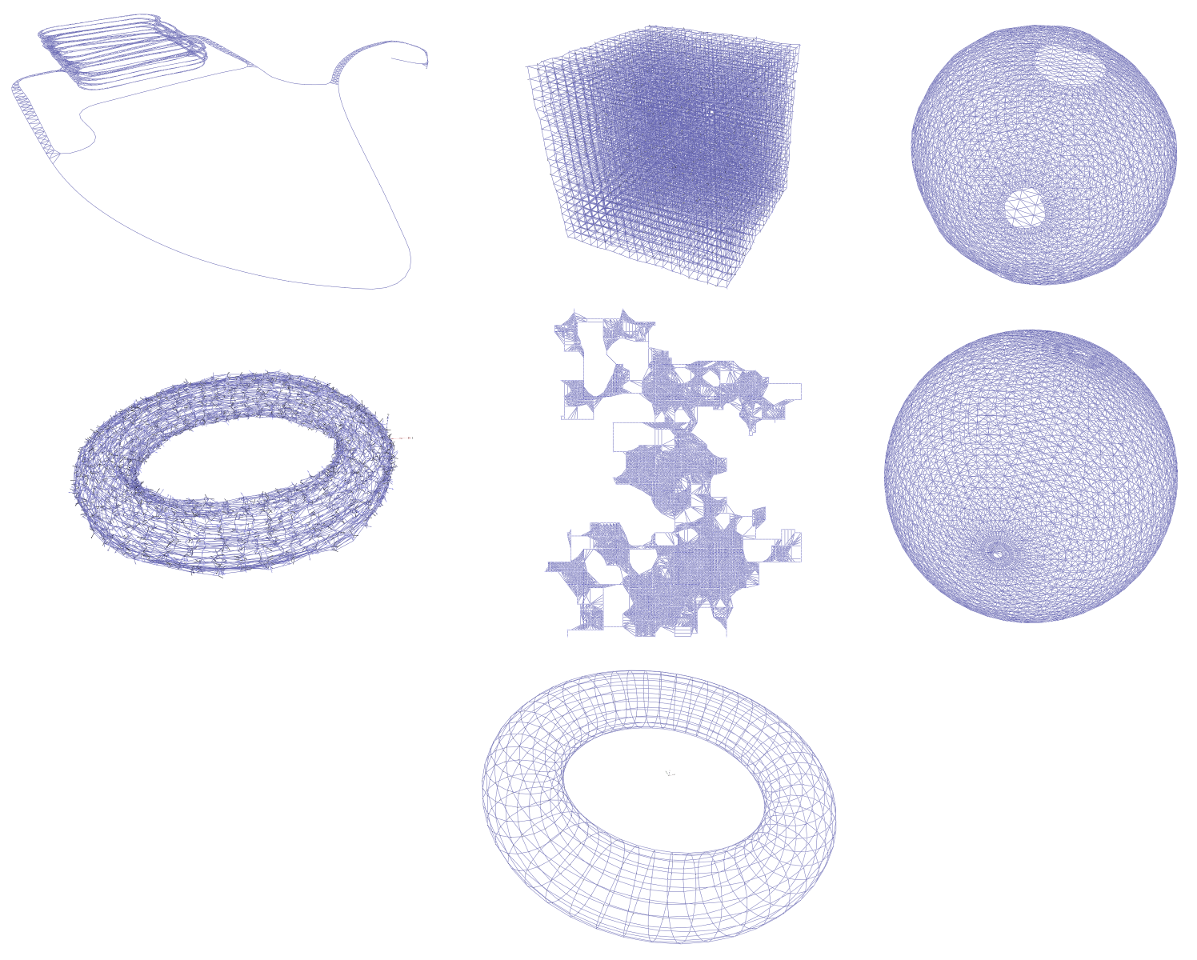}
  \caption{Datasets used to perform the experiments. Top row, from left to 
    right: 
    pose graph of the Stanford parking garage (referred as \texttt{garage}), 
    simulated 3D grid (\texttt{grid}), simulated 3D sphere (\texttt{sphere-a}); 
    mid row, left to right: simulated 3D torus (\texttt{torus-a}), simulated 3D 
    dataset (\texttt{sim-manhattan}), simulated 3D sphere (\texttt{sphere-b}); 
    last 
    row: simulated 3D torus (\texttt{torus-b}).}
  \label{fig:datasets}
  \vspace{-15pt}
\end{figure}

\noindent We tested our error function embedding it within the \g2o 
optimization framework. 
Datasets specifications are available in~\figref{fig:datasets} and 
in~\tabref{tab:dataset-specs}.

To evaluate the performances of approaches under varying noise
conditions, we added to the original datasets noise sampled from
$\mathcal{N}_t(0, \Sigma_t)$ and $\mathcal{N}_R(0, \Sigma_R)$
respectively for the translational and rotational component of the
pose.  Then we analyzed the convergence using the chordal and the
geodesic error functions, varying both the statistical parameters of
the noise distributions and the initial guess.  To compare the
residual error evolution between the two error functions, we recompute
the chi2 - i.e. the quadratic error obtained summing the $\be_k$ computed for 
each measurement $\bZ_k$ - at each iteration of the chordal optimization 
using the geodesic function.  
In Section~\ref{par:spherical-covariance} we
present the result obtained with spherical covariances.  In
Section~\ref{par:brvtte-covariance} we report the effects of the
optimization under generic covariances. Since the value of parameter
$\epsilon$ controls the conversion between geodesic and chordal problem, we 
investigated the effects of this parameter in Section~\ref{par:ezzilo}.

\subsection{Spherical Covariances}
\label{par:spherical-covariance}
\begin{figure}[t]
  \centering
  \includegraphics[width=0.99\columnwidth]{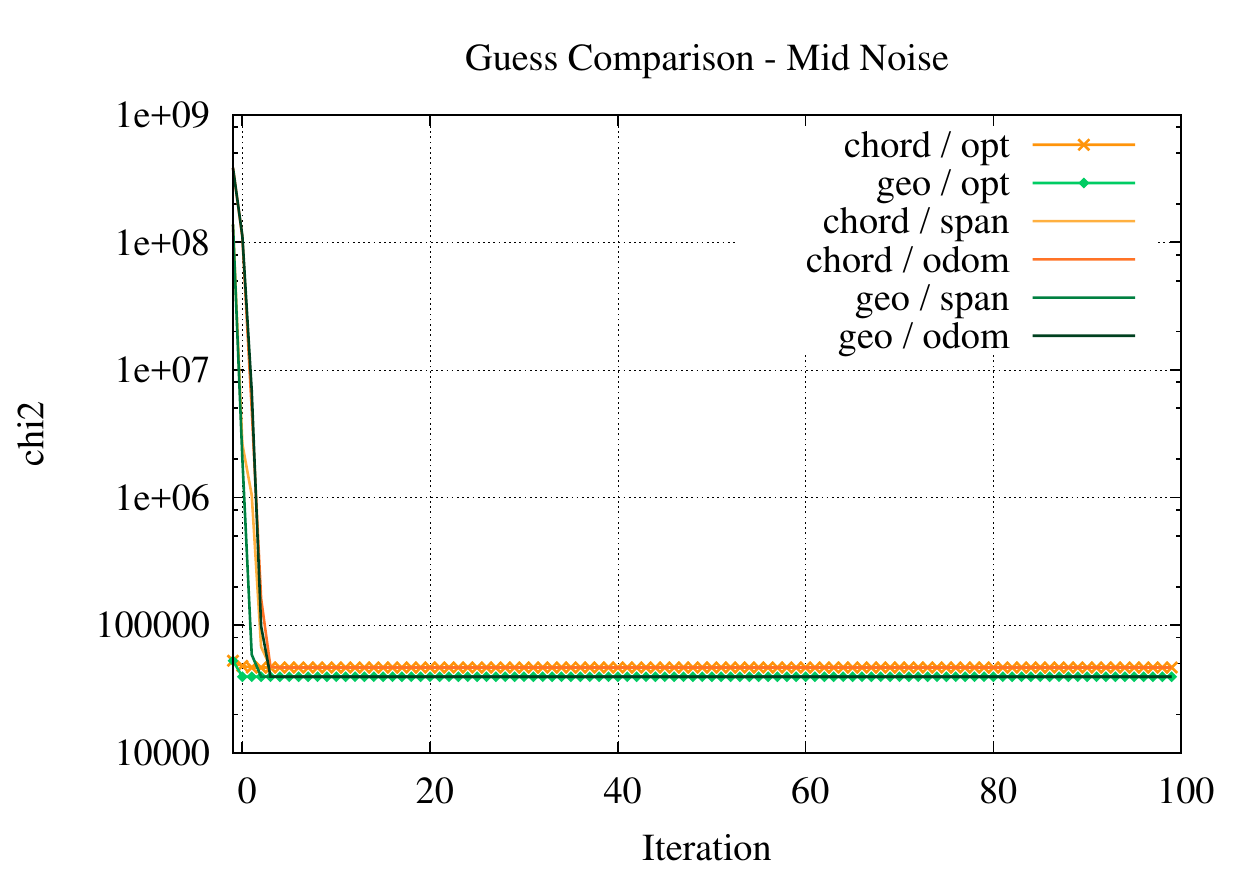}
  \caption{Evolution of the chi2 in dataset \texttt{sphere-a} with noise 
  covariances $\Sigma_t = [0.1\;0.1\;0.1] [m]$ and $\Sigma_R = [0.01\; 0.01\; 
  0.01] [rad]$. Both approaches easily succeed.}
  \label{fig:spehere-a-gn-chi2-mid}
  \vspace{-15pt}
\end{figure}

In the first set of experiments, we added a relatively small noise figure to 
the pose measurements. In particular, the statistical parameters are $\Sigma_t 
= [0.1\;0.1\;0.1] [m]$ and $\Sigma_R = [0.01\; 0.01\; 0.01] [rad]$. As shown 
in~\figref{fig:spehere-a-gn-chi2-mid}, both error function succeed in finding 
the optimum, although our approach requires slightly more iterations.

\begin{figure*}[t]
  \centering
  \begin{subfigure}{0.31\linewidth}
    \includegraphics[width=0.9\linewidth]{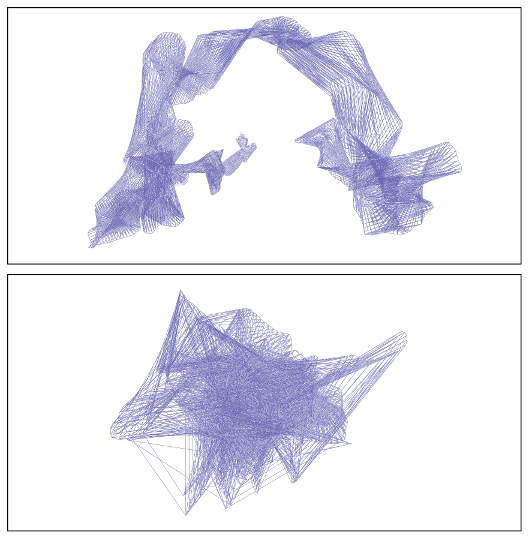}
    \subcaption{Initial guess from the odometry (top) and spanning tree 
      (bottom).}
    \label{fig:sphere-a-gn-initial-guesses}
  \end{subfigure} \hspace{5px}
  \begin{subfigure}{0.31\textwidth}
    \includegraphics[width=0.9\linewidth]{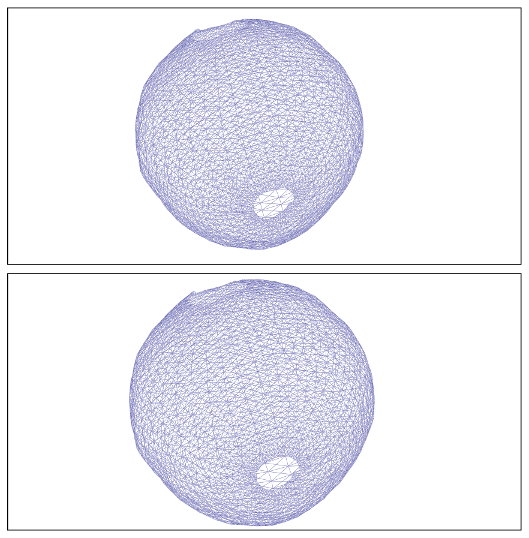}
    \subcaption{GN chordal optimization outputs using odometry (top) and 
      spanning tree (bottom) as initial guess.}
    \label{fig:sphere-a-gn-chordal}
  \end{subfigure}  \hspace{5px}
  \begin{subfigure}{0.31\textwidth}
    \includegraphics[width=0.9\linewidth]{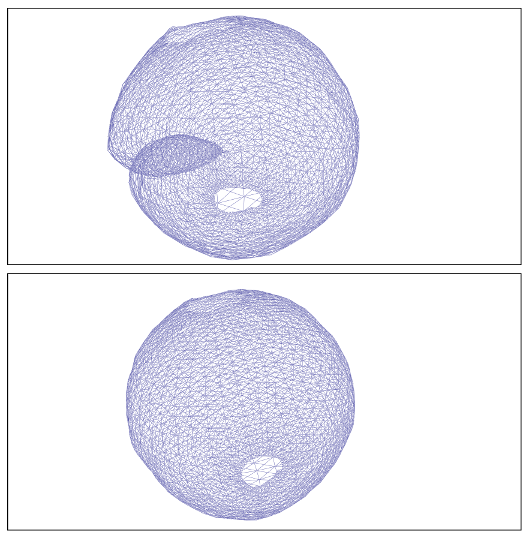}
    \subcaption{GN geodesic optimization outputs using odometry (top) and 
      spanning tree (bottom) as initial guess.}
    \label{fig:sphere-a-gn-geodesic}
  \end{subfigure} \\ 
  \begin{subfigure}{0.48\textwidth}
    \includegraphics[width=0.99\linewidth]{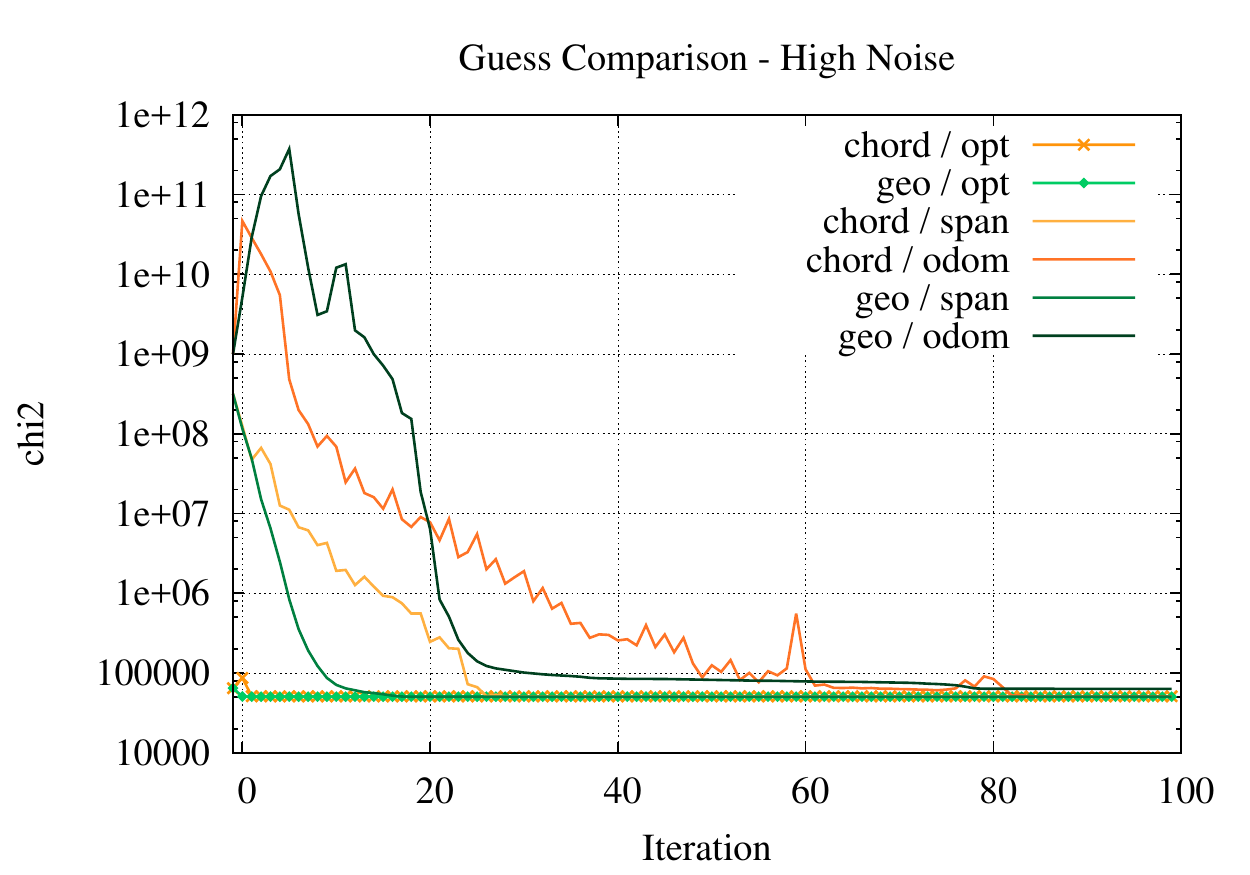}
    \subcaption{Comparison of the chi2 using Gauss-Newton as optimization 
      algorithm.}
    \label{fig:spehere-a-gn-chi2}
  \end{subfigure} \hspace{5px}
  \begin{subfigure}{0.48\textwidth}
    \includegraphics[width=0.99\linewidth]{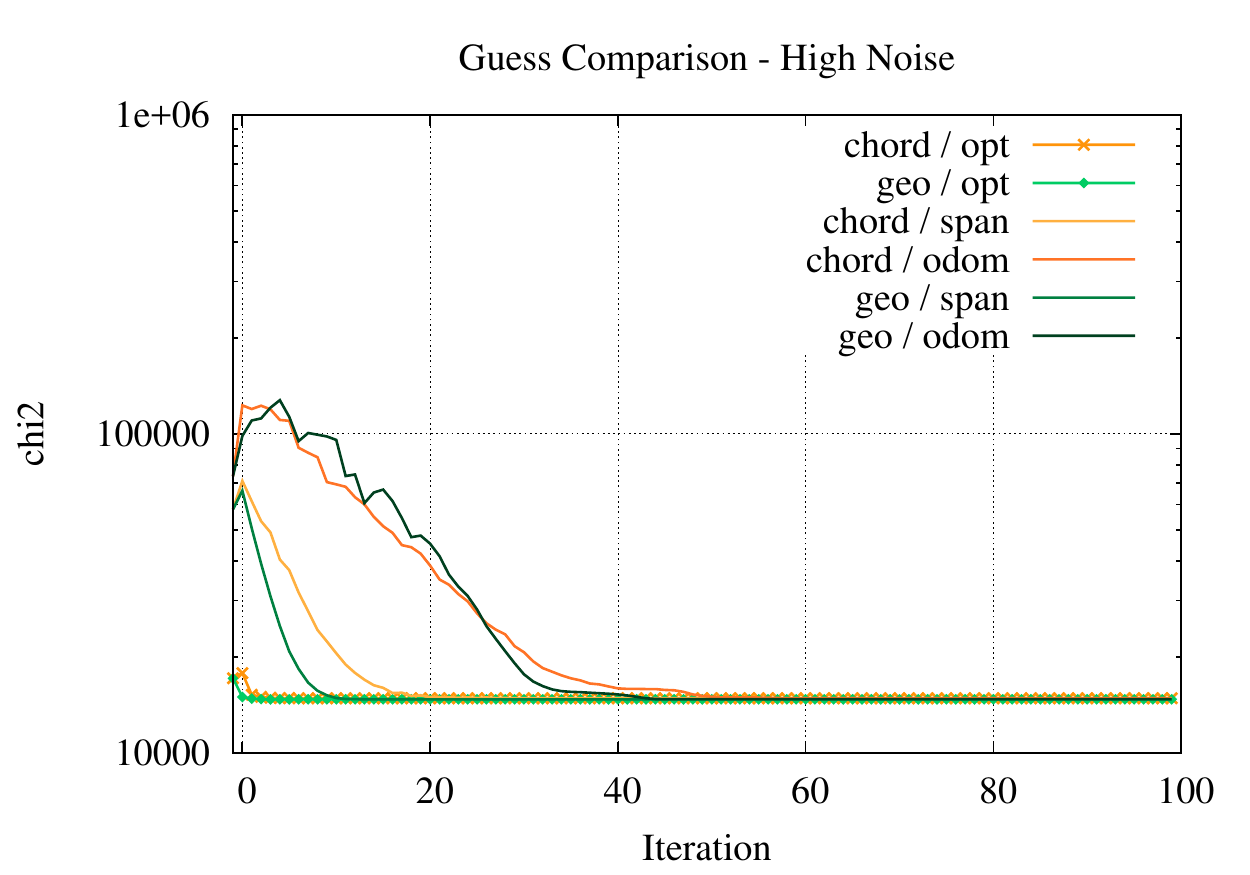}
    \subcaption{Comparison of the chi2 using Gauss-Newton and a Cauchy kernel 
      with width $k = 1.0$.}
    \label{fig:sphere-a-gn-k-chi2}
  \end{subfigure} 
  \caption{Analysis of the \texttt{sphere-a} synthetic dataset. We encoded in 
    the measurements noise sampled from Gaussian distributions with $\Sigma_t = 
    [0.5 \; 0.5 \; 0.5] [m]$ and $\Sigma_R = [0.1 \; 0.1 \; 0.1] [rad]$. The 
    initial guesses computed from the odometry traversal and the spanning tree 
    are reported in~\figref{fig:sphere-a-gn-initial-guesses}. 
    The output produced by the Gauss-Newton (GN) optimization using the chordal 
    and the geodesic distances are illustrated respectively 
    in~\figref{fig:sphere-a-gn-chordal} and~\figref{fig:sphere-a-gn-geodesic}. 
    Even in this case the geodesic distance remains stuck in a local minimum, 
    as highlighted in~\figref{fig:spehere-a-gn-chi2}. 
    Introducing a Cauchy kernel in the optimization process, both functions 
    reach the same optimum as shown in~\figref{fig:sphere-a-gn-k-chi2}.}
  \label{fig:sphere-a-all-noise}
  \vspace{-15pt}
\end{figure*}

Then, we increased the noise using $\Sigma_t = [0.5 \; 0.5 \; 0.5] [m]$ 
and $\Sigma_R = [0.1 \; 0.1 \; 0.1] [rad]$. In this case the noise components 
are very high in each pose dimension. Results of the 
optimization process on the \texttt{sphere-a} dataset are shown 
in~\figref{fig:sphere-a-all-noise}. 
These initial guesses are extremely poor, and neither of the two approaches can reach the optimum.
However, the chordal function produces better results with respect to the 
geodesic one. 

\subsection{Non-Spherical Covariances}
\label{par:brvtte-covariance}
\begin{figure*}[t]
  \centering
  \begin{subfigure}{0.31\linewidth}
    \includegraphics[width=0.9\linewidth]{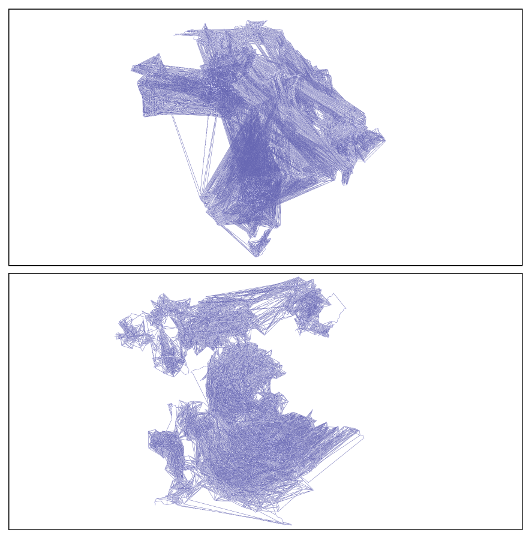}
    \subcaption{Top: initial guess from odometry; bottom: initial guess 
      from spanning tree}
    \label{fig:sim-5000p-gn-initial-guess}
  \end{subfigure} \hspace{5px}
  \begin{subfigure}{0.31\textwidth}
    \includegraphics[width=0.9\linewidth]{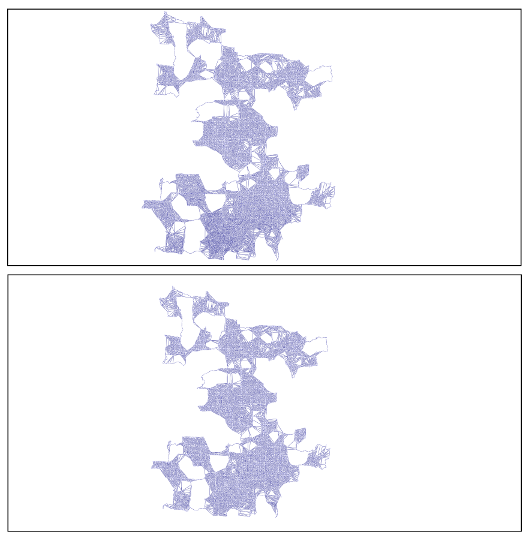}
    \subcaption{GN chordal output from odometry (top) and from spanning tree 
      (bottom).}
    \label{fig:sim-5000p-gn-chordal}
  \end{subfigure}  \hspace{5px}
  \begin{subfigure}{0.31\textwidth}
    \includegraphics[width=0.9\linewidth]{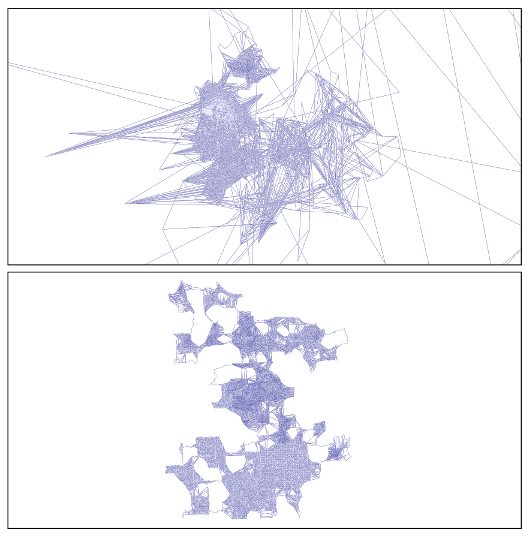}
    \subcaption{GN geodesic outputs from odometry (top) and from spanning 
    tree (bottom).}
    \label{fig:sim-5000p-gn-geodesic}
  \end{subfigure} \\
  \begin{subfigure}{0.48\textwidth}
    \includegraphics[width=0.99\linewidth]{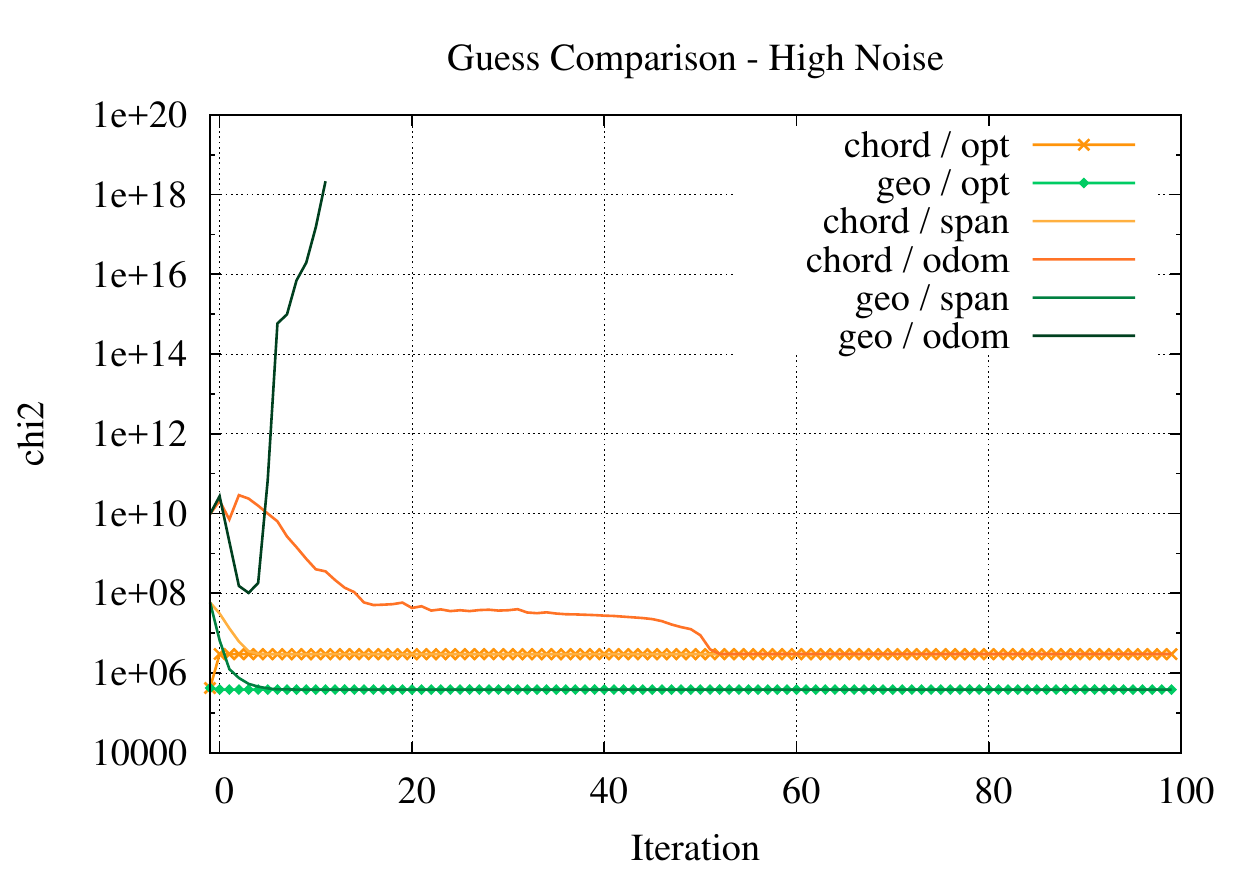}
    \subcaption{Comparison of the chi2 using Gauss-Newton as optimization 
      algorithm. If the initial guess is far from the optimum - e.g. odometry 
      case - the geodesic error function stops the optimization before the end 
      of 
      the iterations due to numerical problems.}
    \label{fig:sim-5000p-gn-chi2}
  \end{subfigure} \hspace{5pt}
  \begin{subfigure}{0.48\textwidth}
    \includegraphics[width=0.99\linewidth]{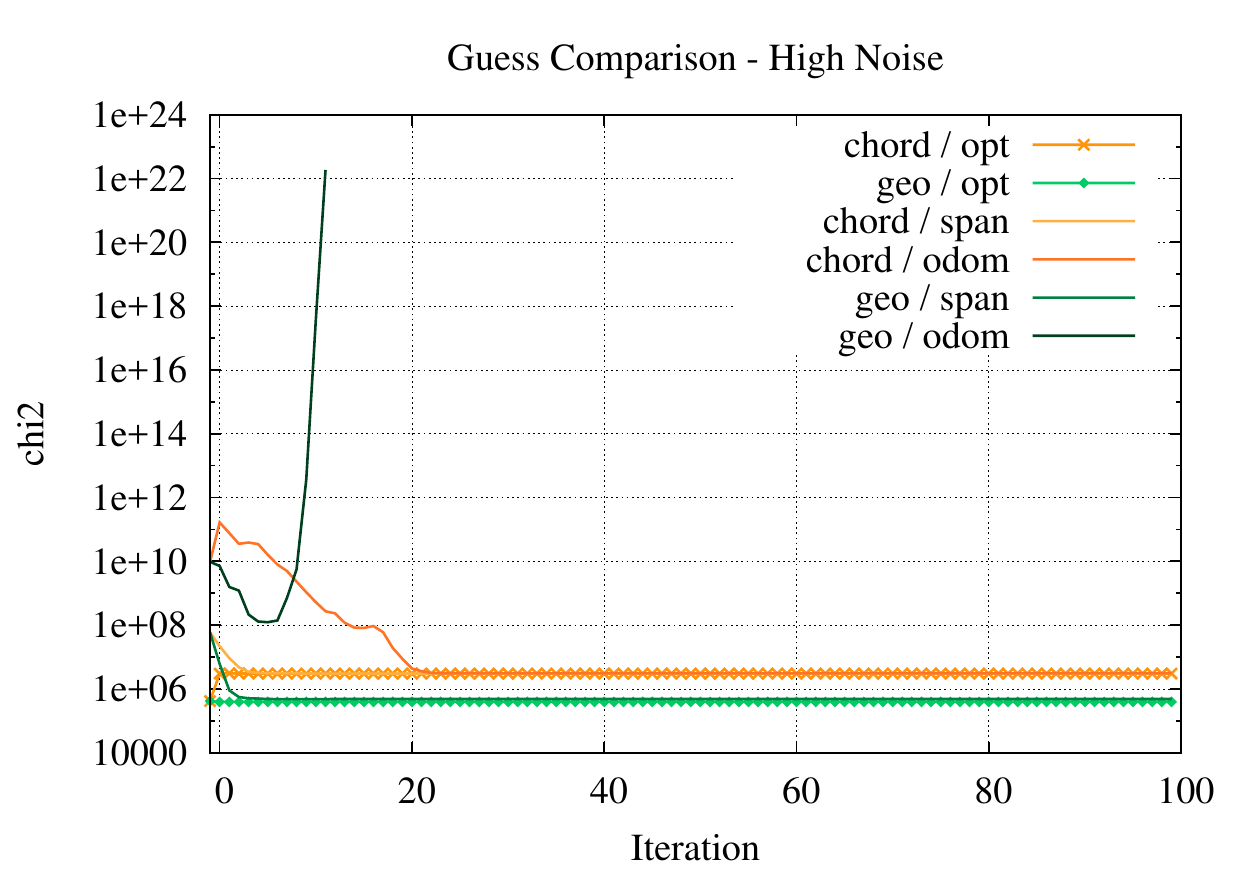}
    \subcaption{Comparison of the chi2 using Gauss-Newton and a Cauchy kernel 
      with kernel width $k = 1.0$. When the initial guess is far from 
      the optimum, the optimization stops before the end of the 
      iterations using the geodesic distance.}
    \label{fig:sim-5000p-gn-k-chi2}
  \end{subfigure}
  \caption{Analysis of the \texttt{sim-manhattan} dataset. We encoded in the 
    measurements noise sampled from Gaussian distributions with $\Sigma_t = 
    [0.5 \; 0.5 \; 0.01] [m]$ and $\Sigma_R = [0.0001 \; 0.0001 \; 0.1] [rad]$. 
    With such noise, the initial guesses computed through odometry and spanning 
    tree are  reported in~\figref{fig:sim-5000p-gn-initial-guess}. 
    The results after 100 iterations of Gauss-Newton (GN) optimization using 
    the chordal and the geodesic error function are depicted respectively  
    in~\figref{fig:sim-5000p-gn-chordal} 
    and~\figref{fig:sim-5000p-gn-geodesic}. 
    Using the standard error function, the 
    optimization process will fail, due to numeric issue - i.e. Hessian 
    non-PSD. \figref{fig:sim-5000p-gn-chi2} illustrates the residual error - 
    i.e. the $\chi^2$ - evolution over 100 iterations. Adding a Cauchy robust 
    kernel to the geodesic optimization, the final state reached is not as far 
    from the optimum as in the previous case, however the problem is still 
    numerically unstable. 
    Curves marked as \texttt{opt} in the legend show the optimization evolution 
    using the optimum as initial guess, so they are used as reference for the 
    two error functions.}
  \label{fig:sim-5000p-gn}
  \vspace{-15pt}
\end{figure*}

For the second set of experiments, 
we used $\Sigma_t = [0.5 \; 0.5 \; 0.01] [m]$ 
and $\Sigma_R = [0.0001 \; 0.0001 \; 0.1] [rad]$. With this noise figures, we 
want to investigate the effects of extremely non-spherical measurements 
covariance matrices in the optimization process. 
In this configuration our error function can reach the optimum 
even when the geodesic error function remains stuck in a local minimum or the 
linear system cannot be solved due to numerical issues. 
In~\figref{fig:sim-5000p-gn} the reader can find the analysis 
of such case for the \texttt{sim-mahattan} dataset. The results obtained from 
the other datasets are consistent with \texttt{sim-mahattan}, and we omit them
for sake of brevity.

We observed that when the rotational noise is particularly large - e.g. 
$\Sigma_R = [0.1\;0.1\;0.1]$ - using the geodesic error function for the 
optimization leads to solutions that are further from the optimum than the ones 
reported by our approach.
Intuitively, large values of rotational noise tend to excite more the 
non-linearities in the error function, that are the main source of 
non-convexity.

In conclusion we observed that the proposed error function exhibits a larger
convergence basin compared to the geodesic one while requiring a 
slightly higher number of iterations in order to reach the optimum.

\subsection{Influence of Covariance Conversion on the Optimum}
\label{par:ezzilo}
The reader might notice that in~\figref{fig:sim-5000p-gn-chi2}, the 
two approaches converge to a slightly different optimum. This mismatch is due 
to the value $\epsilon$ used to convert the information matrix. 
The results reported for all experiments are obtained by using a value 
of $\epsilon=0.1$. Using such a value has negligible effects 
on the minimum of the converted error problem when the measurements are 
affected by a standard deviation in the same order of magnitude as $\epsilon$.
In this case, the optimum of the geodesic and the chordal problems are 
equivalent in terms of chi2 - e.g. as reported 
in~\figref{fig:spehere-a-gn-chi2} and~\figref{fig:sphere-a-gn-k-chi2}.
However, when $\epsilon$ is large compared to the noise in one or more 
dimensions, the two optima are in slightly different, albeit visual 
inspection of the pose graph reveal no substantial inconsistencies.
This problem can be approached in two alternative ways:
\begin{itemize}
  \item[--] Start an optimization using the geodesic error function from the 
  optimum obtained by using the chordal function. In general the chordal 
  solution represents a very good starting point, and the geodesic error 
  function converges in a few steps.
  \item[--] Dynamically adapt the value of $\epsilon$ using an adaptive 
  strategy based on the rate of convergence. This results in a strategy similar
  to the Levenberg-Marquardt algorithm.
\end{itemize}

\begin{figure}
  \centering
  \includegraphics[width=0.98\columnwidth]{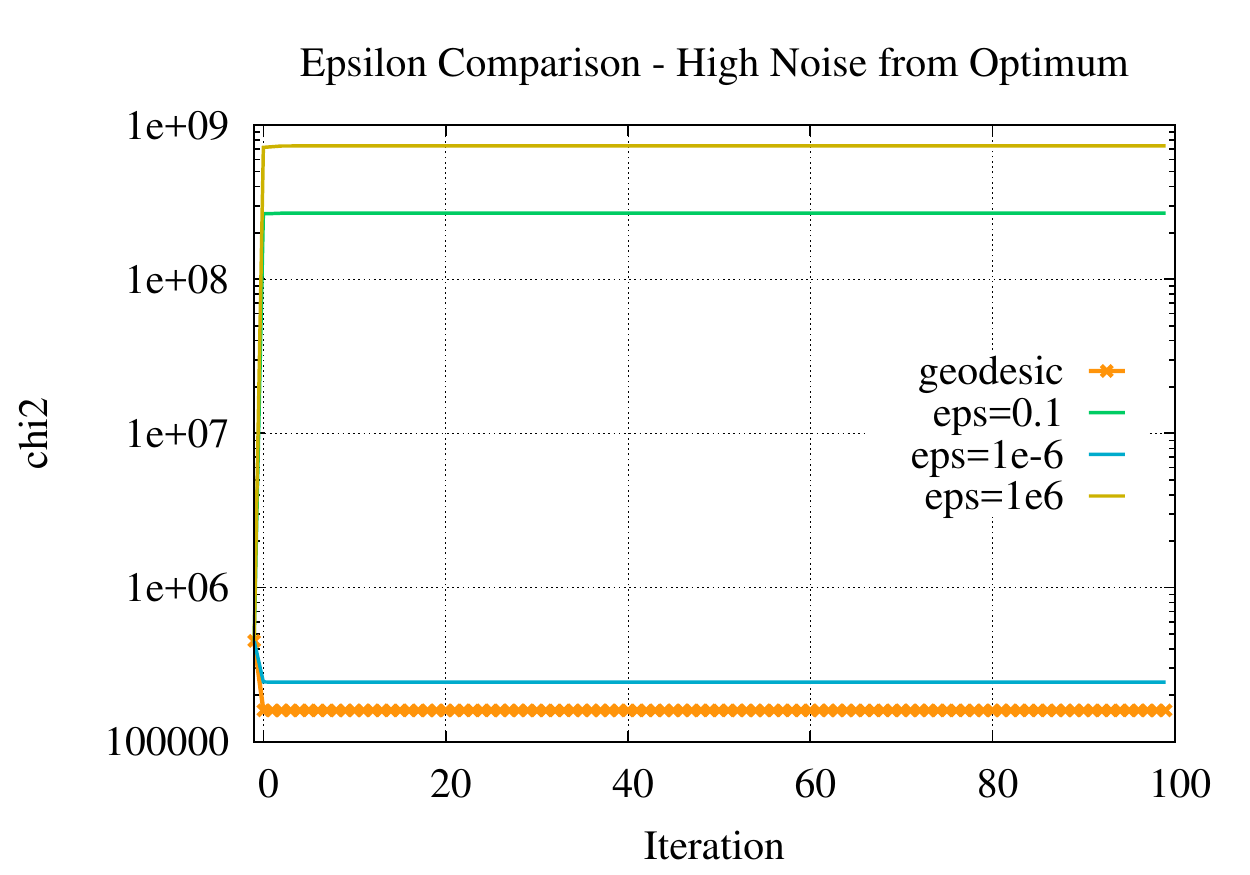}
  \caption{Evolution of the chi2 on the \texttt{sphere-a} dataset with noise 
  statistics equal to $\Sigma_t = [0.5 \; 0.5 \; 0.01] [m]$ and $\Sigma_R = 
  [0.0001 \; 0.0001 \; 0.1] [rad]$. The smaller $\epsilon$, the closer the 
  chordal optimum to the geodesic one.}
  \label{fig:epsilon-comparison}
\end{figure}

To characterize the influence of parameter $\epsilon$ in the optimization, 
we performed a third experiment . We perturbed the 
\texttt{sphere-a} dataset adding the following noise figures
$\Sigma_t = [0.5 \; 0.5 \; 0.01] [m]$ and 
$\Sigma_R = [0.0001 \; 0.0001 \; 0.1] [rad]$. 
In~\figref{fig:epsilon-comparison} we reported the chi2 of different 
optimizations obtained varying the value of $\epsilon$, using the optimum as 
initial guess. The experiments confirm our conjectures that $\epsilon$ values 
larger than the noise standard deviation - on one or more dimensions - will 
smoothen the error surface leading to different optima with respect to the one 
retrieved 
with the geodesic distance.

\section{Conclusions}\label{sec:conclusions}
In this work we proposed an alternative error function for 3D pose-graph 
optimization problems, based on the chordal distance between matrices rather 
than the geodesic one. Its main features are: 
(i)~reduction of problem's non-linearities with a consequent enlarged 
convergence basin and a greater robustness to rotational noise,
(ii)~derivatives easy to compute in close form and that lead to nice numerical 
properties of the Jacobians - e.g. one is the opposite of the other - and, 
thus, to a faster computation of matrix $\textbf{H}$.

Our conjunctures are confirmed by a large set of comparative experiments.
To use the chordal error function, one has to convert the 
problem expressed in geodesic form. Under realistic conditions, the 
converted problem has a solution equivalent to the original one.
Under extremely uneven noise figures the two optima might be different,
however we by visual inspection we were not able to spot inconsistencies 
in the returned solutions.


%
%
\appendices
\section{Jacobians' Computation}\label{sec:appendix}
In this small Appendix we provide the mathematical derivation of the Jacobians both in the standard parametrization and in the chordal one.
\paragraph{Standard Formalization} 
Let~$\bX$ be a 3D-isometry composed as in~\eqref{eq:extended-parametrization}; let~$\bDeltax$ be a 6-vector defined as in~\eqref{eq:minimal-parametrization}. The functions v2t and t2v map a 6-vector into a 3D-isometry and vice-versa. The former one ensembles the transformation as follows:
\begin{align}
  \text{v2t}(\bDeltax) &= \begin{pmatrix}\bR & \bt \\ \bZero_{3 \times 1} & 1\end{pmatrix} \quad \bt = \begin{bmatrix}\Delta x & \Delta y & \Delta z\end{bmatrix}^T \label{eq:v2t}\\
  \bR &= \bR_x(\Delta\phi)\,\bR_y(\Delta\theta)\,\bR_z(\Delta\psi) \nonumber
\end{align} 
where $\bR_x$,~$\bR_y$ and~$\bR_z$ are the standard 3D rotation matrices around the respective axis. As a result, indicating with~$c$ the $\cos$ and with~$s$ the~$\sin$ of an angle, matrix~$\bR$ is computed as:
\begin{align}
  \label{eq:rot-composition-xyz}
  \bR &= 
  \begin{bmatrix}
  R_{00} & R_{01} & R_{02} \\
  R_{10} & R_{11} & R_{12} \\
  R_{20} & R_{21} & R_{22} 
  \end{bmatrix}
  = \bR_x(\Delta\phi) \, \bR_y(\Delta\theta)\, \bR_z(\Delta\psi) = \\
  &=
  \begin{bmatrix}
  c\Delta\theta\,c\Delta\psi & {}\;{} & -c\Delta\theta\,s\Delta\psi & {}\;{} & s\Delta\theta \\
  m & {}\;{} & n & {}\;{} & -c\Delta\theta\,s\Delta\phi \\
  p & {}\;{} & q & {}\;{} & c\Delta\theta\,c\Delta\phi 
  \end{bmatrix} \nonumber
\end{align}
where 
\begin{align*}
  m &= c\Delta\phi\,s\Delta\psi + s\Delta\phi\,c\Delta\psi\,s\Delta\theta\\
  n &= c\Delta\phi\,c\Delta\psi - s\Delta\phi\,s\Delta\theta\,s\Delta\psi\\
  p &= s\Delta\phi\,s\Delta\psi - c\Delta\phi\,c\Delta\psi\,s\Delta\theta\\
  q &= s\Delta\phi\,c\Delta\psi + c\Delta\phi\,s\Delta\theta\,s\Delta\psi
\end{align*}
Given this, with the function t2v we have to perform the inverse process, retrieving the Euler angles~$\Delta \phi$, $\Delta \theta$ and $\Delta \psi$ from~\eqref{eq:rot-composition-xyz}. As a consequence of this, the Jacobians $\tilde \bJ_i$ and $\tilde \bJ_j$ computed through~\eqref{eq:jac-i-increments} and~\eqref{eq:jac-j-increments} are really complex and full of non-linear components.

\paragraph{Alternative Formalization}
In this case, we do not use the t2v function in the~$\boxminus$, but the difference between two isometries is computed according to~\eqref{eq:boxminus-chordal}. Given the error function in~\eqref{eq:error-chordal}, applying a small state perturbation~$\bDeltax$, it will become:
\begin{align}
  \nonumber
  {}&\be_{ij}(\bX_i \boxplus \bDeltax_i, \bX_j \boxplus \bDeltax_j) = \\ 
  &= \flatten{\left(\text{v2t}(\bDeltax_i)\bX_i\right)^{-1} \, \left(\text{v2t}(\bDeltax_j)\bX_j\right)} - \flatten{\bZ_{ij}}
  \label{eq:perturbed-error-chordal}
\end{align}
The Jacobian $\tilde \bJ_j$ is computed performing the partial derivative of~\eqref{eq:perturbed-error-chordal} w.r.t.~$\bDeltax_j$:
\begin{equation}
  \tilde \bJ_j = \frac{\partial\: \be_{ij}(\bX_i \boxplus \bDeltax_i, \bX_j \boxplus \bDeltax_j)}{\partial \bDeltax_j} \Bigg \rvert_{\scriptsize\begin{matrix} \bDeltax_i = 0\\\bDeltax_j = 0\end{matrix}}
  \label{eq:jac-j-appendix}
\end{equation}
Therefore, we define the following matrices:
\begin{align}
  \bA &= \begin{bmatrix}\bR_i^T & -\bR_i^T \bt_i \\\bZero & 1\end{bmatrix} \\
  \bB &= \begin{bmatrix}\left(\bR_{\bDeltax_i}^x \, \bR_{\bDeltax_i}^y \, \bR_{\bDeltax_i}^z\right)^T & -\bR_{\bDeltax_i}^T \bt_i \\\bZero & 1\end{bmatrix} \\
  \bC &= \begin{bmatrix}\bR_j & \bt_j \\\bZero & 1\end{bmatrix}
\end{align}
where:
\begin{itemize}
  \item  $\bR_{x0}^{\prime}$, $\bR_{y0}^{\prime}$ and $\bR_{z0}^{\prime}$ that represent derivatives with respect to $\Delta\phi$, $\Delta\theta$ and $\Delta\psi$ of the base rotation $\bR_k(\cdot)$, evaluated in 0 and with $k = \{x,y,z\}$;
  \item  $\hat{\bR}_{x0}^{\prime}$, $\hat{\bR}_{y0}^{\prime}$, and $\hat{\bR}_{z0}^{\prime}$ that are the derivatives with respect to $\Delta\phi$, $\Delta\theta$ and $\Delta\psi$ of the rotational part of matrix $\mathbf{G}$, computed as $\hat{\bR}_{k0}^{\prime} = \bR_{i}^T \, \bR_{k0}^{\prime} \, \bR_j$ with $k = \{x,y,z\}$;
\end{itemize}
We indicate with $\hat{\mathbf{r}}_{k0}^{\prime}$ the 9 vector obtained stacking the columns of $\hat{\bR}_{k0}^{\prime}$ - with $k = \{x,y,z\}$, and, as a result, the Jacobian becomes:
\begin{align}
  \tilde \bJ_j &= \frac{\partial \left[\flatten{\bA \bB \bC}\right]}{\partial \bDeltax_j} \Bigg \rvert_{\scriptsize\begin{matrix} \bDeltax_i = 0\\\bDeltax_j = 0\end{matrix}} \nonumber \\
  &= 
  \begin{pmatrix}
    \bZero_{(9\times3)} & \begin{bmatrix} \hat\br_{x0}^{\prime} & | & \hat\br_{y0}^{\prime} & | & \hat\br_{z0}^{\prime} \end{bmatrix}_{(9\times3)} \\ 
    \bR_{i}^T & - \bR_{i}^T \, \skew{\bt_j}
  \end{pmatrix}
  \label{eq:jac-j-chordal} 
\end{align}
Finally, $\tilde \bJ_i$ can be computed straightforwardly 
from~\eqref{eq:jac-j-chordal}, leading to the relation
\begin{equation*}
  \tilde \bJ_i = -\tilde \bJ_j
\end{equation*}

\bibliographystyle{plain}
\bibliography{robots}

\begin{thebibliography}{10}

\bibitem{ceres-solver}
Sameer Agarwal, Keir Mierle, and Others.
\newblock Ceres solver.
\newblock http://ceres-solver.org.

\bibitem{Carlone11rss-lagoPGO2D}
L.~Carlone, R.~Aragues, J.A. Castellanos, and B.~Bona.
\newblock A linear approximation for graph-based simultaneous localization and
  mapping.
\newblock In {\em Proc. of Robotics: Science and Systems (RSS)}, pages 41--48,
  2011.

\bibitem{carlone2015initialization}
Luca Carlone, Roberto Tron, Kostas Daniilidis, and Frank Dellaert.
\newblock Initialization techniques for 3d slam: a survey on rotation
  estimation and its use in pose graph optimization.
\newblock In {\em Robotics and Automation (ICRA), 2015 IEEE International
  Conference on}, pages 4597--4604. IEEE, 2015.

\bibitem{dellaert2012gtsam}
Frank Dellaert.
\newblock Factor graphs and gtsam: A hands-on introduction.
\newblock Technical report, Georgia Institute of Technology, 2012.

\bibitem{dellaert2006square}
Frank Dellaert and Michael Kaess.
\newblock Square root sam: Simultaneous localization and mapping via square
  root information smoothing.
\newblock {\em The International Journal of Robotics Research},
  25(12):1181--1203, 2006.

\bibitem{Duckett02Mapping}
T.~{Duckett}, S.~{Marsland}, and J.~{Shapiro}.
\newblock Fast, on-line learning of globally consistent maps.
\newblock {\em Autonomous Robots}, 12(3):287 -- 300, 2002.

\bibitem{Frese04relaxation}
U.~Frese, P.~Larsson, and T.~Duckett.
\newblock A multilevel relaxation algorithm for simultaneous localisation and
  mapping.
\newblock {\em IEEE Transactions on Robotics}, 21(2):1--12, 2005.

\bibitem{grisetti12iros}
G.~Grisetti, R.~K{\"u}mmerle, and K.~Ni.
\newblock Robust optimization of factor graphs by using condensed measurements.
\newblock In {\em Proc. of the {IEEE/RSJ} Int. Conf. on Intelligent Robots and
  Systems (IROS)}, Vilamoura, Portugal, October 2012.

\bibitem{grisetti2010tutorial}
Giorgio Grisetti, Rainer Kummerle, Cyrill Stachniss, and Wolfram Burgard.
\newblock A tutorial on graph-based slam.
\newblock {\em IEEE Intelligent Transportation Systems Magazine}, 2(4):31--43,
  2010.

\bibitem{grisetti2007tree}
Giorgio Grisetti, Cyrill Stachniss, Slawomir Grzonka, and Wolfram Burgard.
\newblock A tree parameterization for efficiently computing maximum likelihood
  maps using gradient descent.
\newblock In {\em Robotics: Science and Systems}, volume~3, page~9, 2007.

\bibitem{gutmann1999incremental}
J-S Gutmann and Kurt Konolige.
\newblock Incremental mapping of large cyclic environments.
\newblock In {\em Computational Intelligence in Robotics and Automation, 1999.
  CIRA'99. Proceedings. 1999 IEEE International Symposium on}, pages 318--325.
  IEEE, 1999.

\bibitem{hertzberg2012tutorial}
Christoph Hertzberg, Ren{\'e} Wagner, and Udo Frese.
\newblock Tutorial on quick and easy model fitting using the slom framework.
\newblock In {\em International Conference on Spatial Cognition}, pages
  128--142. Springer, 2012.

\bibitem{Howard2001relaxation}
A.~Howard, M.J. Matari\'{c}, and G.~Sukhatme.
\newblock Relaxation on a mesh: a formalism for generalized localization.
\newblock In {\em Proc.~of the IEEE/RSJ Int.~Conf.~on Intelligent Robots and
  Systems (IROS)}, 2001.

\bibitem{kaess2012isam2}
Michael Kaess, Hordur Johannsson, Richard Roberts, Viorela Ila, John~J Leonard,
  and Frank Dellaert.
\newblock isam2: Incremental smoothing and mapping using the bayes tree.
\newblock {\em The International Journal of Robotics Research}, 31(2):216--235,
  2012.

\bibitem{kaess2007isam}
Michael Kaess, Ananth Ranganathan, and Frank Dellaert.
\newblock isam: Fast incremental smoothing and mapping with efficient data
  association.
\newblock In {\em Robotics and Automation, 2007 IEEE International Conference
  on}, pages 1670--1677. IEEE, 2007.

\bibitem{kummerle2011g}
Rainer K{\"u}mmerle, Giorgio Grisetti, Hauke Strasdat, Kurt Konolige, and
  Wolfram Burgard.
\newblock g 2 o: A general framework for graph optimization.
\newblock In {\em Robotics and Automation (ICRA), 2011 IEEE International
  Conference on}, pages 3607--3613. IEEE, 2011.

\bibitem{lu1997globally}
Feng Lu and Evangelos Milios.
\newblock Globally consistent range scan alignment for environment mapping.
\newblock {\em Autonomous robots}, 4(4):333--349, 1997.

\bibitem{NiICRA2007}
Kai Ni, Drew Steedly, and Frank Dellaert.
\newblock Tectonic sam: Exact, out-of-core, submap-based slam.
\newblock In {\em Proc.~of the IEEE Int.~Conf.~on Robotics \& Automation
  (ICRA)}, 2007.

\bibitem{olson2006fast}
Edwin Olson, John Leonard, and Seth Teller.
\newblock Fast iterative alignment of pose graphs with poor initial estimates.
\newblock In {\em Robotics and Automation, 2006. ICRA 2006. Proceedings 2006
  IEEE International Conference on}, pages 2262--2269. IEEE, 2006.

\bibitem{Smi90EstABBREV}
R.~Smith, M.~Self, and P.~Cheeseman.
\newblock Estimating uncertain spatial realtionships in robotics.
\newblock In I.~Cox and G.~Wilfong, editors, {\em Autonomous Robot Vehicles},
  pages 167--193. Springer Verlag, 1990.

\end{thebibliography}

\end{document}